\documentclass[11pt,letterpaper]{article}
\usepackage{url}
\usepackage[hyperref]{ijcnlp2017}
\usepackage{times}
\usepackage{latexsym}
\usepackage{amsmath}
\usepackage{amsfonts}
\usepackage[small]{caption}
\usepackage{adjustbox}
\usepackage{microtype}
\usepackage{inconsolata}

\def\ijcnlppaperid{1015}

\ijcnlpfinalcopy

\usepackage{todonotes}

\usepackage{cleveref}
\crefname{section}{\S}{\S\S}
\Crefname{section}{\S}{\S\S}
\crefname{table}{Tab.}{}
\crefname{table*}{Tab.}{}
\crefname{figure}{Fig.}{}
\crefname{algorithm}{Alg.}{}
\crefname{appendix}{App.}{}
\crefname{equation*}{Eq.}{}
\crefformat{section}{\S#2#1#3}  

\usepackage{booktabs}

\newcommand{\citeposs}[1]{\citeauthor{#1}'s \citeyearpar{#1}}

\newcommand{\ttt}{\mathbf{t}}
\newcommand{\ww}{\mathbf{w}}
\newcommand{\ee}{\mathbf{e}}
\newcommand{\oo}{\mathbf{o}}
\newcommand{\sss}{\mathbf{s}}
\newcommand{\lll}{\mathbf{l}}
\newcommand{\vtheta}{{\boldsymbol \theta}}
\newcommand{\vomega}{{\boldsymbol \omega}}
\newcommand{\veta}{{\boldsymbol \eta}}
\newcommand{\vf}{{\boldsymbol f}}
\newcommand{\PF}{Z_{\vtheta}(\ww)}
\newcommand{\potent}{\psi\left(t_{i-1}, t_i, \ww ; \vtheta\right)}

\title{Low-Resource Named Entity Recognition with Cross-Lingual, Character-Level Neural Conditional Random Fields}

\author{Ryan Cotterell \quad Kevin Duh \\
  Department of Computer Science \\
  Johns Hopkins University \\  Baltimore, MD 21218 \\
  {\tt
\href{mailto:ryan.cotterell@jhu.edu}{\texttt{ryan.cotterell@jhu.edu}}}}

\date{}

\begin{document}
\sloppy
\maketitle

\begin{abstract}
  Low-resource named entity recognition is still an open problem in
  NLP.  Most state-of-the-art systems require tens of thousands of
  annotated sentences in order to obtain high performance. However,
  for most of the world's languages, it is unfeasible to obtain such
  annotation. In this paper, we present a transfer learning scheme,
  whereby we train character-level neural CRFs to predict named
  entities for both high-resource languages and low-resource languages
  jointly.  Learning character representations for multiple related
  languages allows transfer among the languages, improving $F_1$ by up to 9.8 points
  over a log-linear CRF baseline.
\end{abstract}

\section{Introduction}
Named entity recognition (NER) presents a challenge for modern machine
learning, wherein a learner must deduce which word tokens refer to
people, locations and organizations (along with other possible entity
types). The task demands that the learner generalize {\em from} limited
training data and {\em to} novel entities, often in new
domains. Traditionally, state-of-the-art NER models have relied on
hand-crafted features that pick up on distributional cues as well as
portions of the word forms themselves. In the past few years, however,
neural approaches that jointly learn their own features have surpassed
the feature-based approaches in performance. Despite their empirical
success, neural networks have remarkably high sample complexity and
still only outperform hand-engineered feature approaches when enough
supervised training data is available, leaving effective training of
neural networks in the low-resource case a challenge. 

For most of the world's languages, there is a very limited amount of
training data for NER; CoNLL---the standard dataset in the
field---only provides annotations for 4 languages
\cite{sang2002introduction,TjongKimSang-DeMeulder:2003:CONLL}. Creating
similarly sized datasets for other languages has a prohibitive
annotation cost, making the low-resource case an important
scenario. To get around this barrier, we develop a cross-lingual
solution: given a low-resource target language, we additionally offer
large amounts of annotated data in a language that is genetically
related to the target language. We show empirically that this improves the
quality of the resulting model.

In terms of neural modeling, we introduce a novel neural conditional random
field (CRF) for cross-lingual NER that allows for cross-lingual
transfer by extracting character-level features
using recurrent neural networks, shared among multiple languages;
this tying of parameters enables cross-lingual abstraction.
With experiments on 15 languages, we confirm that feature-based CRFs
outperform the neural methods consistently in the low-resource
training scenario. However, with the addition of cross-lingual
information, the tables turn and the neural methods are again
on top, demonstrating that cross-lingual supervision is a viable method
to reduce the training data state-of-the-art neural
approaches require.

\begin{figure}
  \centering
 \includegraphics[width=\columnwidth]{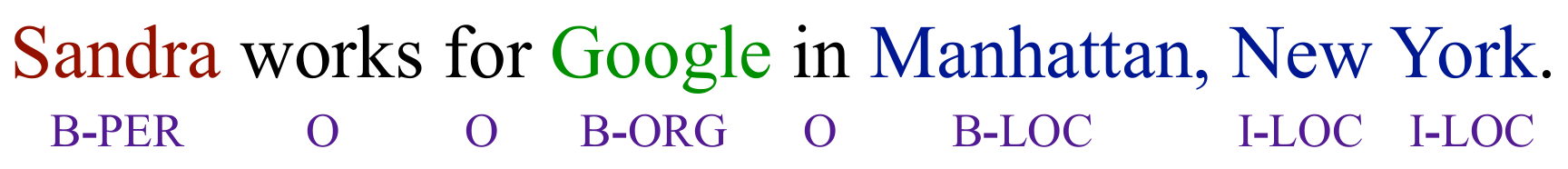}
  \caption{Example of an English sentence annotated with its typed named entities.}
  \label{fig:sentence}
  \vspace{-7.5pt}
\end{figure}

\section{Neural Conditional Random Fields}
Named entity recognition is typically
framed as a sequence labeling
task using the {\sc bio} scheme
\cite{ramshaw1995text, bos}, i.e., given an input sentence, the goal is to
assign a label to each token: {\sc b} if the token is the {\bf b}eginning of
an entity, or {\sc i} if the token is {\bf i}nside an entity, or {\sc o} if
the token is {\bf o}utside an entity (see \cref{fig:sentence}).
Following convention, we focus on person (per), location (loc), organization (org), and miscellaneous (misc) entity types, 
resulting in 9 tags:  \{{\sc b-org}, {\sc i-org}, {\sc
  b-per}, {\sc i-per}, {\sc b-loc}, {\sc i-loc}, {\sc b-misc}, {\sc
  i-misc}\}.

Conditional random fields (CRFs), first introduced in
\newcite{DBLP:conf/icml/LaffertyMP01}, generalize the classical
maximum entropy models \cite{berger1996maximum} to distributions over
structured objects, and are an effective tool for sequence labeling tasks like NER.
We briefly overview the formalism here and then discuss its neural parameterization.
\subsection{CRFs: A Cursory Overview}
We start with two discrete alphabets $\Sigma$ and $\Delta$.  In the
case of sentence-level sequence tagging, $\Sigma$ is a set
of words (potentially infinite) and $\Delta$ is a set of
tags (generally finite; in our case $|\Delta|=9$). 
Given $\ttt = t_1\cdots t_n \in \Delta^n$ and $\ww =
w_1 \cdots w_n \in \Sigma^n$, where $n$ is the sentence length,
a CRF is a globally normalized conditional probability
distribution,
\begin{equation}
  p_{\vtheta}(\ttt \mid \ww) = \frac{1}{\PF} \prod_{i=1}^{n} \potent,
\end{equation}
where $t_0$ is a distinguished beginning-of-tagging symbol, $\potent \geq 0$ is an arbitrary non-negative
potential function\footnote{We slightly abuse notation and use $t_0$
  as a distinguished beginning-of-tagging symbol.} that we take to be
a parametric function of the parameters $\vtheta$ and
the partition function $\PF$ is the sum over all taggings of length $n$.

So how do we choose $\potent$? We discuss two
alternatives, which we will compare experimentally in \cref{sec:experiments}.

\subsection{Log-Linear Parameterization}\label{sec:log-linear}
Traditionally, computational linguists
have considered a simple log-linear parameterization, i.e.,
\begin{align}
\!\potent =\exp(\veta^{\top}\vf(t_{i-1}, t_i, \ww, i))
\end{align}
where $\veta \in \mathbb{R}^d$ and the user defines a feature function
$\vf \colon \Delta \times \Delta \times \Sigma^n \times \mathbb{N} \to \mathbb{R}^d$ that
extracts relevant information from the adjacent tags $t_{i-1}$ and
$t_{i}$ and the sentence $\ww$ at position $i$. In this case, the model's parameters
are $\vtheta = \{ \veta \}$. Common binary features include word form
features, e.g., does the word at the $i^\text{th}$ position end in
{\em -ation}?, and contextual features, e.g., is the word next to
$(i\!-\!1)^{\text{th}}$ word {\em the}? These binary features are
conjoined with other indicator features, e.g., is the $i^\text{th}$
tag {\sc i-loc}? We refer the reader to \newcite{ShaP03} for standard
CRF feature functions employed in NER, which we use in this work.
The log-linear parameterization
yields a convex objective and is extremely efficient to compute as it
only involves a sparse dot product, but the representational power of
the model depends fully on the quality of the features the user selects.

\subsection{(Recurrent) Neural Parameterization}\label{sec:log-nonlinear}
Modern CRFs, however, try to obviate the hand-selection of features
through deep, non-linear parameterizations of $\potent$. This idea is
far from novel and there have been numerous attempts in the literature
over the past decade to find effective non-linear parameterizations
\cite{PengBX09,artieres2010neural,collobert2011natural,vinel2011joint,fujii2012deep}. Until recently, however, it was not clear that these non-linear
parameterizations of CRFs were worth the non-convexity and the extra
computational cost.  Indeed, on neural CRFs,
\newcite{wang-manning:2013:IJCNLP2} find that ``a nonlinear
architecture offers no benefits in a high-dimensional discrete feature
space.''

However, recently with the application of long short-term memory
\cite[LSTM;][]{HochreiterS97} recurrent neural networks
\cite[RNNs;][]{Elman90} to CRFs, it has become clear that neural feature
extractors are superior to the hand-crafted approaches
\cite{huang2015bidirectional,lample-EtAl:2016:N16-1,P16-1101}. As our starting
point, we build upon the architecture of
\newcite{lample-EtAl:2016:N16-1}, which is currently competitive with
the state of the art for NER.
\begin{align}
  \psi(t_{i-1}, t_i, &\ww ; \left.\vtheta\right)   = \\
  &\exp(a(t_{i-1}, t_i)+\oo(t_i)^{\top} \mathbf{W}\,\sss(\ww)_i), \nonumber
\end{align}
where $a(t_{i-1}, t_i)$ is the weight of transitioning from $t_{i-1}$ to $t_i$
and $\oo(t_i) \in \mathbb{R}^{r_1}$, $\sss(\ww)_i \in \mathbb{R}^{r_2}$ and $\mathbf{W} \in \mathbb{R}^{r_1 \times r_2}$ are the output tag representation, the contextual representation of the $i^\text{th}$ word, and an interaction matrix, respectively. We define the contextual representation $\sss(\ww)_i$ of the $i^\text{th}$ word as the concatenation of the forward and backward hidden states, at position $i$, of an LSTM run over the sentence,\footnote{We take $r_1 = r_2 = 100$ and use a two-layer LSTM with $200$ hidden units each. The sentence-level $\text{LSTM}_{\vtheta'}$ of Eq.~\ref{eq:sentence-representation} returns its full sequence of hidden states---one per word---so that $\sss(\ww)_i$ is its output at position $i$; the character-level $\text{LSTM}_{\vtheta''}$ of Eq.~\ref{eq:word-representation} instead returns only its final hidden state, summarizing the word.} i.e.,
\begin{equation}
\sss(\ww)_i = \left[ \overrightarrow{\text{LSTM}_{\vtheta'}}(\vomega)_i;\, \overleftarrow{\text{LSTM}_{\vtheta'}}(\vomega)_i\right].
    \label{eq:sentence-representation}
\end{equation}
The input vector $\vomega = [\omega_1, \ldots, \omega_{|\ww|}]$ to this BiLSTM is
a vector of representations: we define
\begin{equation}
  \omega_i = \left[\text{LSTM}_{\vtheta''}\left(c_1 \cdots
    c_{|w_i|}\right);\, \ee(w_i)\right],
    \label{eq:word-representation}
\end{equation}
where $c_1\cdots c_{|w_i|}$ are the characters in word $w_i$. In other words, we run an
LSTM over the character stream and concatenate it with a word
representation for each type. Now, the parameters $\vtheta$, $\vtheta'$ and $\vtheta''$ are the set of all the LSTM parameters
and other representations. 

\section{Cross-Lingual Extensions}
One of the most striking features of neural networks is their ability to
abstract general representations across many words. 
Our question is: can neural feature-extractors abstract the notion of
a named entity across similar languages?  For example, if we train a
character-level neural CRF on several of the highly related Romance
languages, can our network learn a general representation of the entities in these languages?

\subsection{Cross-Lingual Architecture}
We now describe our novel cross-lingual architecture. Given a language
label $\ell$, we want to create a language-specific CRF $p_{\vtheta}(\ttt
\mid \ww, \ell)$ with potential:
\begin{align}
  \psi(t_{i-1},t_i, &\ww, \ell ; \vtheta) = \exp\left(a(t_{i-1}, t_i)+\right. \label{eq:crosslingual} \\
  & \mathbf{u}(t_i)^{\top}\!\tanh(\mathbf{U} \left.\left[\right.\sss(\ww)_i;\, \lll(\ell)\left.\right] + \mathbf{b})\right), \nonumber
\end{align}
$\lll(\ell) \in \mathbb{R}^{r_3}$ is a representation of the language
ID itself, $\mathbf{U} \in \mathbb{R}^{q \times (r_2 + r_3)}$ is a projection matrix, $\mathbf{b} \in \mathbb{R}^q$ is a bias vector, $\mathbf{u}(t_i) \in \mathbb{R}^q$ is a tag-specific output vector, and $a(\cdot, \cdot)$ is a transition weight for each pair of tags. 
As the distribution is now conditioned on the language $\ell$, the corresponding partition function becomes $Z_{\vtheta}(\ww, \ell)$, summing $\prod_{i=1}^{n} \psi(t_{i-1}, t_i, \ww, \ell; \vtheta)$ over all taggings of $\ww$.
Importantly, we share some parameters across languages: the
transitions $a()$ between tags and the character-level neural
networks that discover what a form looks like. 
Recall $\sss(\ww)_i$ is defined in Eq.~\ref{eq:sentence-representation} and Eq.~\ref{eq:word-representation}.
The character encoder $\text{LSTM}_{\vtheta''}\left(c_1 \cdots c_{|w_i|}\right)$ is shared cross-lingually while $\ee(w_i)$ is language-specific.
All remaining parameters are likewise shared across languages---the sentence BiLSTM $\vtheta'$, the projection $\mathbf{U}$, the tag-specific output vectors $\mathbf{u}(t_i)$ and the bias $\mathbf{b}$---so that the only language-specific components are the word representations $\ee(w_i)$ and the language representation $\lll(\ell)$, the latter modulating the shared network for each language. The potential of Eq.~\eqref{eq:crosslingual} defines the neural CRF used in all of our experiments; in the monolingual setting, with no source language, the model still carries a single language representation, so the monolingual and transfer neural models share one architecture and differ only in their training data.

Now, given a low-resource target language $\tau$ and a source language
$\sigma$ (potentially, a set of $m$ high-resource source languages
$\{\sigma_i\}_{i=1}^m$), we consider the following log-likelihood objective
\begin{align}
  {\cal L}\left(\vtheta\right) &= \!\! \sum_{(\ttt, \ww) \in {\cal
      D}_{\tau}} \!\!\!\!\! \log p_{\vtheta}\left(\ttt \mid \ww, \tau
  \right) + \\ &
  \hspace{2cm} \mu \cdot \!\!\!\!\!\! \sum_{(\ttt, \ww) \in {\cal D}_{\sigma}} \!\!\!\!\!
  \log p_{\vtheta}\left(\ttt \mid \ww, \sigma \right), \nonumber
\end{align}
where $\mu > 0$ is a trade-off parameter, ${\cal D}_{\tau}$ is the set of training examples for the
target language and ${\cal D}_{\sigma}$ is the set of training data
for the source language $\sigma$. In the case of multiple source
languages, we add a summand for each source language, so that we have multiple training sets ${\cal
  D}_{\sigma_i}$.

In the case of the log-linear parameterization, we simply add a
language-specific atomic feature for the language-id, drawing
inspiration from \citeposs{daumeiii:2007:ACLMain} approach for domain
adaptation. We then conjoin this new atomic feature with the existing
feature templates, doubling the number of feature templates: the
original and the new feature template conjoined with
the language ID.
\begin{table}
  \centering
  \begin{adjustbox}{width=.80\columnwidth}
  \begin{tabular}{llll} \toprule
    Language     & Code & Family          & Branch \\ \midrule
    Galician     & gl   & Indo-European   & Romance \\
    Catalan      & ca   & Indo-European   & Romance \\
    French       & fr   & Indo-European   & Romance \\
    Italian      & it   & Indo-European   & Romance \\
    Romanian     & ro   & Indo-European   & Romance \\
    Spanish      & es   & Indo-European   & Romance \\ \midrule
    West Frisian & fy   & Indo-European   & Germanic \\
    Dutch        & nl   & Indo-European   & Germanic  \\ \midrule
    Tagalog      & tl   & Austronesian    & Philippine \\
    Cebuano      & ceb  & Austronesian    & Philippine \\ \midrule
    Ukrainian     & uk   & Indo-European   & Slavic \\
    Russian      & ru   & Indo-European   & Slavic \\ \midrule
    Marathi      & mr   & Indo-European   & Indo-Aryan \\
    Hindi        & hi   & Indo-European   & Indo-Aryan \\
    Urdu         & ur   & Indo-European   & Indo-Aryan \\
    \bottomrule
  \end{tabular}
  \end{adjustbox}
  \caption{List of the languages used in our experiments with their ISO 639-1 codes, family and the branch in that family.}
  \label{tab:languages}
\end{table}

\begin{table*}
  \centering
  \begin{tabular}{ll | ccc ccc} \toprule
    & languages                  & \multicolumn{3}{c}{low-resource ($|{\cal D}_{\tau}|=100$)} & \multicolumn{3}{c}{high-resource ($|{\cal D}_{\tau}|=10000$)} \\ \cmidrule(l){1-2} \cmidrule(l){3-5} \cmidrule(l){6-8}
    $\tau$ & $\sigma_i$  & log-linear    & neural     & ${\boldsymbol \Delta}$ & log-linear   & neural    & ${\boldsymbol \Delta}$ \\ \midrule
    gl  & ---            & $57.64$       & $49.19$    & $\textcolor{red}{-8.45}$  & $87.23$      & $89.42$   & $\textcolor{blue}{+2.19}$ \\ 
    gl  & es             & $71.46$       & $76.40$    & $\textcolor{blue}{+4.94}$  & $87.50$      & $89.46$   & $\textcolor{blue}{+1.96}$ \\
    gl  & ca             & $67.32$       & $75.40$    & $\textcolor{blue}{+8.08}$  & $87.40$      & $89.32$   & $\textcolor{blue}{+1.92}$ \\
    gl  & it             & $63.81$       & $70.93$    & $\textcolor{blue}{+7.12}$  & $87.34$      & $89.50$   & $\textcolor{blue}{+2.16}$ \\
    gl  & fr             & $58.22$       & $68.02$    & $\textcolor{blue}{+9.80}$  & $87.92$      & $89.38$   & $\textcolor{blue}{+1.46}$ \\
    gl  & ro             & $59.32$       & $67.76$    & $\textcolor{blue}{+8.44}$  & $87.24$      & $89.19$   & $\textcolor{blue}{+1.95}$ \\ \midrule
    fy  & ---            & $62.71$       & $58.43$    & $\textcolor{red}{-4.28}$  & $90.42$      & $91.03$   & $\textcolor{blue}{+0.61}$ \\
    fy  & nl             & $68.15$       & $72.12$    & $\textcolor{blue}{+3.97}$  & $90.94$      & $91.01$   & $\textcolor{blue}{+0.07}$ \\ \midrule
    tl  & ---            & $58.15$       & $56.98$    & $\textcolor{red}{-1.17}$  & $74.24$      & $79.03$   & $\textcolor{blue}{+4.79}$ \\
    tl  & ceb            & $75.29$       & $81.79$    & $\textcolor{blue}{+6.50}$  & $74.02$      & $79.51$   & $\textcolor{blue}{+5.48}$ \\ \midrule
    uk  & ---            & $61.40$       & $60.65$    & $\textcolor{red}{-0.75}$  & $85.63$      & $87.39$   & $\textcolor{blue}{+1.75}$ \\
    uk  & ru             & $70.94$       & $76.74$    & $\textcolor{blue}{+5.80}$  & $86.01$      & $87.42$   & $\textcolor{blue}{+1.41}$ \\ \midrule
    mr  & ---            & $42.76$       & $39.02$    & $\textcolor{red}{-3.73}$  & $70.98$      & $74.95$   & $\textcolor{blue}{+3.97}$ \\
    mr  & hi             & $54.25$       & $60.92$    & $\textcolor{blue}{+6.67}$  & $70.45$      & $74.49$   & $\textcolor{blue}{+4.04}$ \\
    mr  & ur             & $49.32$       & $58.92$    & $\textcolor{blue}{+9.60}$  & $70.75$      & $74.81$   & $\textcolor{blue}{+4.07}$ \\ \bottomrule
  \end{tabular}
  \caption{Results comparing the log-linear and neural CRFs in various settings. We compare the log-linear and the neural CRF in the low-resource
    transfer setting. The difference (${\boldsymbol \Delta}$) is blue when positive and red when negative.}
  \label{tab:results}
  \vspace{-7.5pt}
\end{table*}

\section{Related Work}
We divide the discussion of related work topically.

\paragraph{Character-Level Neural Networks.}
In recent years, many authors have incorporated character-level
information into taggers using neural networks, e.g.,
\newcite{dos2014learning} employed a convolutional network for
part-of-speech tagging in morphologically rich languages and
\newcite{ling-EtAl:2015:EMNLP2} an LSTM for a myriad of different
tasks.  Relatedly, \newcite{Q16-1026} approached NER with
character-level LSTMs, but without using a CRF. Our work firmly builds
upon this in that we, too, compactly summarize the
word form with a recurrent neural component.

\paragraph{Neural Transfer Schemes.}
Previous work has also performed transfer learning using neural networks. The
novelty of our work lies in the {\em cross-lingual} transfer. For
example, \newcite{peng-dredze-2017} and \newcite{yang2017transfer},
similarly oriented concurrent papers, focus on domain adaptation
within the same language. While this is a related problem,
cross-lingual transfer is much more involved since the morphology,
syntax and semantics change more radically between two languages than between domains.

\paragraph{Projection-based Transfer Schemes.}
Projection is a common approach to tag low-resource languages.
The strategy involves annotating one side of bitext
with a tagger for a high-resource language and then projecting the
annotations over the bilingual alignments obtained through
unsupervised learning \cite{och2003systematic}.  Using these projected
annotations as weak supervision, one then trains a tagger in the
target language. This line of research has a rich history, starting
with \newcite{yarowsky2001inducing}.  For a recent take, see
\newcite{wang2014cross} for projecting NER from English to Chinese.
We emphasize that projection-based approaches are {\em incomparable to our proposed method}
as they make an additional bitext assumption, which is generally not present in the case
of low-resource languages.

\section{Experiments}\label{sec:experiments}

Fundamentally, we want to show that character-level neural CRFs are
capable of generalizing the notion of an entity across related
languages. To get at this, we compare a linear CRF (see
\cref{sec:log-linear}) with standard feature templates for the task and
a neural CRF (see \cref{sec:log-nonlinear}). We further compare three
training set-ups: low-resource, high-resource and low-resource with
additional cross-lingual data for transfer. Given past results in the
literature, we expect linear CRF to dominate in the low-resource
settings, the neural CRF to dominate in the high-resource setting. The
novelty of our paper lies in the consideration of the low-resource
with transfer case: we show that neural CRFs are better at transferring
entity-level abstractions cross-linguistically.

\subsection{Data}
We experiment on 15 languages from the cross-lingual named entity
dataset described in \newcite{hengji:2017:P17}. We focus on 5 typologically diverse\footnote{While most of
  these languages are from the Indo-European family, they still run
  the gauntlet along a number of typological axes, e.g., Dutch and
  Ukrainian and the Indo-Aryan languages employ postpositions
  (attached to the word) rather than prepositions (space-separated).}
target languages: Galician, West Frisian, Ukrainian, Marathi and
Tagalog.  As related source languages, we consider Spanish, Catalan,
Italian, French, Romanian, Dutch, Russian, Cebuano, Hindi and
Urdu. For the language code abbreviations and linguistic families, see
\cref{tab:languages}. For each of the target languages, we emulate a truly low-resource condition, creating a 100-sentence split
for training. We then create a 10000-sentence superset
to be able to compare to a high-resource condition in those same languages. For the source
languages, we only created a 10000-sentence split. We also create disjoint validation and test splits of 1000
sentences each.

\subsection{Results}
The linear CRF is trained using L-BFGS \cite{Liu_1989} until convergence using the
CRF suite toolkit.\footnote{{\tiny
    \url{http://www.chokkan.org/software/crfsuite/}}} We train our
neural CRF for 100 epochs using {\sc AdaDelta}
\cite{zeiler2012adadelta} with a learning rate of $1.0$. The results
are reported in \cref{tab:results}. To understand the table, take the
target language ($\tau$) Galician. In terms of $F_1$, while the neural
CRF outperforms the log-linear CRF in the high-resource setting ($89.42$
vs. $87.23$), it performs poorly in the low-resource setting ($49.19$
vs. $57.64$); when we add in a source language ($\sigma_i$)
such as Spanish, $F_1$ increases to $76.40$ for the neural CRF and
$71.46$ for the log-linear CRF. The trend is similar for other source
languages, such as Catalan ($75.40$) and Italian ($70.93$).\looseness=-1

Overall, we observe three general trends. i) In the monolingual
high-resource case, the neural CRF outperforms the log-linear CRF. ii)
In the low-resource case, the log-linear CRF outperforms the neural
CRF. iii) In the transfer case, the neural CRF wins, however,
indicating that our character-level neural approach is truly better at
generalizing cross-linguistically in the low-resource case (when we
have little target language data), as we hoped. In the high-resource
case (when we have a lot of target language data), the transfer
learning has little to no effect. We conclude that our cross-lingual
neural CRF is a viable method for the transfer
of NER. However, there is still a sizable gap between the neural
CRF trained on 10000 target sentences and the transfer case (100
target and 10000 source), indicating there is still room for
improvement.

\section{Conclusion}
We have investigated the task of cross-lingual transfer in
low-resource named entity recognition using neural CRFs with
experiments on 15 typologically diverse languages. Overall, we show that direct cross-lingual transfer is an option for
reducing sample complexity for state-of-the-art architectures.
In the future, we plan to investigate how exactly the networks manage to induce a cross-lingual entity abstraction.

\section*{Acknowledgments}
We are grateful to Heng Ji and Xiaoman Pan for sharing their dataset and providing support.

\sloppy
\bibliography{crosslingual-ner}
\bibliographystyle{acl_natbib}

\end{document}